\def\year{2021}\relax
\documentclass[letterpaper]{article} 
\usepackage{aaai21}  
\usepackage{times}  
\usepackage{helvet} 
\usepackage{courier}  
\usepackage[hyphens]{url}  
\usepackage{graphicx} 
\usepackage{natbib}  
\usepackage{caption} 
\title{A Comparison of Multi-View Learning Strategies for Satellite Image-Based Real Estate Appraisal}
\author{
    Jan-Peter Kucklick,\textsuperscript{\rm 1}
    Oliver Müller, \textsuperscript{\rm 1}
    \\
}
\affiliations{
    \textsuperscript{\rm 1}Paderborn University\\

    Warburger Str. 100 \\
    33098, Paderborn, Germany\\
    jan.kucklick@upb.de, oliver.mueller@upb.de

}

\begin{document}
\maketitle

\begin{abstract}
In the house credit process, banks and lenders rely on a fast and accurate estimation of a real estate price to determine the maximum loan value. Real estate appraisal is often based on relational data, capturing the hard facts of the property. Yet, models benefit strongly from including image data, capturing additional soft factors. The combination of the different data types requires a multi-view learning method. Therefore, the question arises which strengths and weaknesses different multi-view learning strategies have. In our study, we test multi-kernel learning, multi-view concatenation and multi-view neural networks on real estate data and satellite images from Asheville, NC. Our results suggest that multi-view learning increases the predictive performance up to 13\% in MAE. Multi-view neural networks perform best, however result in intransparent black-box models. For users seeking interpretability, hybrid multi-view neural networks or a boosting strategy are a suitable alternative. 
\end{abstract}

\section{Introduction}
Real estate markets are an important part of many economies and account for 3-5\% of the yearly Gross Domestic Product (GDP)  in the U.S. \cite{nahb}. While houses are often bought on credit, 70\% of all mortgages are associated with them \cite{stupak}. One essential process within the market is real estate appraisal, the estimation of the monetary value of a property or land. As the appraised value of a property usually determines the upper limit of a credit a home buyer can expect to get from a financial service provider, a timely and accurate estimation of real estate prices is key for banks and other lenders \cite{liu2018learning}.

Traditionally, real estate appraisal was performed with the help of hedonic pricing models introduced by \citeauthor{lancaster1966new} \citeyearpar{lancaster1966new} and \citeauthor{rosen1974hedonic} \citeyearpar{rosen1974hedonic}. These models use linear regressions to estimate the monetary value of a property based on a number of constituting characteristics typically measured as numerical (e.g., size, age) or categorical (e.g., location, condition) variables. Over the last years, several researchers have extended the classical real estate appraisal models by using Convolutional Neural Networks (CNNs), instead of linear regressions, to incorporate image data into the learning process \cite{law2019take}. Examples include interior and exterior perspectives of houses, as well as street-side and satellite imagery \cite{law2019take, poursaeed2018vision, bency2017beyond, bessinger2016quantifying, liu2018learning, kucklick2020location}. Published empirical results strongly suggest that adding such image data to real estate appraisal models improves their predictive performance. However, combining multiple numerical, categorical, and visual features in one predictive model poses a number of challenges related to the fusion of information from heterogeneous sources \cite{li2018survey}. So far, no standard network architecture has emerged to combine the structured numerical and categorical features on the one hand, and the images on the other hand in order to compute a final prediction. The proposed architectures range from multi-kernel learning to multi-view neural networks, and most researchers do not compare the effect of different approaches to the predictive accuracy of their models. 

Against this background, we empirically evaluate different multi-view learning strategies for combining structured and image data in real estate appraisal models. For our experiments, we use structured data of 32,700 real estates from Asheville, NC, and combine it with satellite images from Bing Maps. Our experimental results show that for the given dataset, using multi-view learning leads to an improvement in predictive accuracy of approximately 13\%. We also find that different multi-view learning strategies vary in their accuracy gains and in their interpretability. Overall, we make three contributions in this article: First, we review different multi-view learning strategies from the literature. Second, we empirically test these strategies in terms of predictive accuracy using a common dataset. Third, we discuss strengths and weaknesses of the different approaches from a theoretical and empirical perspective.

\section{Related Work}
\subsection{Real Estate Appraisal}
Real estate appraisal is typically based on hedonic pricing models. In these models, the overall appraisal value is the weighted sum of the partial value contributions of the constituting characteristics of a real estate \cite{lancaster1966new, rosen1974hedonic}. Statistically, these models are expressed in a linear regression and typical features are related to size (e.g., number of rooms, bathrooms, bedrooms, square feet), condition (e.g., exterior, interior quality), or amenities (e.g., air condition, number of parking spaces, fireplaces) \cite{ligus2016measuring, limsombunchai2004house, helbich2013boosting, hill2018can, park2015using}.

A weakness of linear regression models is that they are bound to use structured data only, that is, variables represented in numerical or categorical form \cite{law2019take}. In other words, traditional linear regression models can only consider hard facts about a real estate. Yet, soft facts - for example related to livability or security - also play a considerable role in the real estate buying and evaluation process \cite{law2019take}. In a typical real estate leaflet, various images portrait this information and real estate experts or buyers can easily combine hard and soft facts to come to an overall judgment. For algorithms, however, understanding visual information and combining it with structured data is still a challenge in several ways. First, data in different representations, e.g., structured attributes and satellite images, need to be harmonized \cite{liu2018learning}. Second, for a holistic model, the different data types need to be combined in a suitable way \cite{liu2018learning}. Third, most statistical and machine learning algorithms can only process data in tabular format (e.g., vectors, matrices), as opposed to image data represented in a cube form, which captures the image height, width, and channel dimensions (rank three tensor). 

Deep neural networks, like CNNs, offer new possibilities to cope with the above challenges. For example, in recent work, CNNs have been used to extract features from property images, which were then fused with structured features and fed into a downstream regression model. Several types of real estate images can be used in such an approach, ranging from interior images to satellite images. Interior images, for example, potentially contain information about a home's luxury level and aesthetics \cite{poursaeed2018vision, naumzik2020one}, whereas exterior images can capture the style and look of the property \cite{bessinger2016quantifying}. In contrast, street-side and satellite images may capture information about the neighborhood and spatial relations, setting the focus apart from the property to a local and global sphere \cite{law2019take, bency2017beyond, kucklick2020location}. As prior research has applied a diverse set of strategies to combine such visual information with standard structured attributes, we review and compare different so-called multi-view learning methods in the next section.

\subsection{Multi-View Learning} 
Multi-view learning describes strategies for learning from various distinct data sources \cite{sun2013survey}. Each source (or view) might contain different complementary information and even different representations of data, for example, relational, image, text, or video data. 
Broadly speaking, these multi-view learning strategies can be divided into two groups: alignment and fusion \cite{li2018survey}.

Multi-view learning through alignment tries to capture the relationships between multiple data sources. It is based on the consensus principle and tries to minimize the disagreement between different data views \cite{xu2013survey}. Let $X_1$ and $X_2$ denote different views and let $f(X_1; W_f)$ and $g(X_2; W_g)$ be embedding functions transforming the views into a multi-view aligned space. Multi-view learning alignment aims to minimize the differences in the embedding output between $f$ and $g$. Therefore, the modeling constraint, which is either distance-based, similarity-based, or correlation-based, is optimized \cite{li2018survey}. In the case of total disagreement of $f$ and $g$, the model error has the upper-bound of the maximum of the individual errors, $maximum(error(f), error(g))$. By achieving agreement, each individual performance of $f$ and $g$ is improved. This optimizes the overall performance of the model. Examples of multi-view alignment methods are different versions of canonical correlation analysis (CCA) \cite{li2018survey} or different co-training algorithms \cite{zhao2017multi}.

Multi-view learning through fusion aims at learning a new joint representation of multiple data sources. It is based on the complementary principle and assumes that each view contains separate complementary information not captured by the other views \cite{xu2013survey}. Exemplary techniques are multi-kernel learning, multi-view concatenation, and different types of multi-view neural networks \cite{li2018survey, xu2013survey}. 

\begin{figure}[!b]
\centering
\includegraphics[width=0.9\columnwidth]{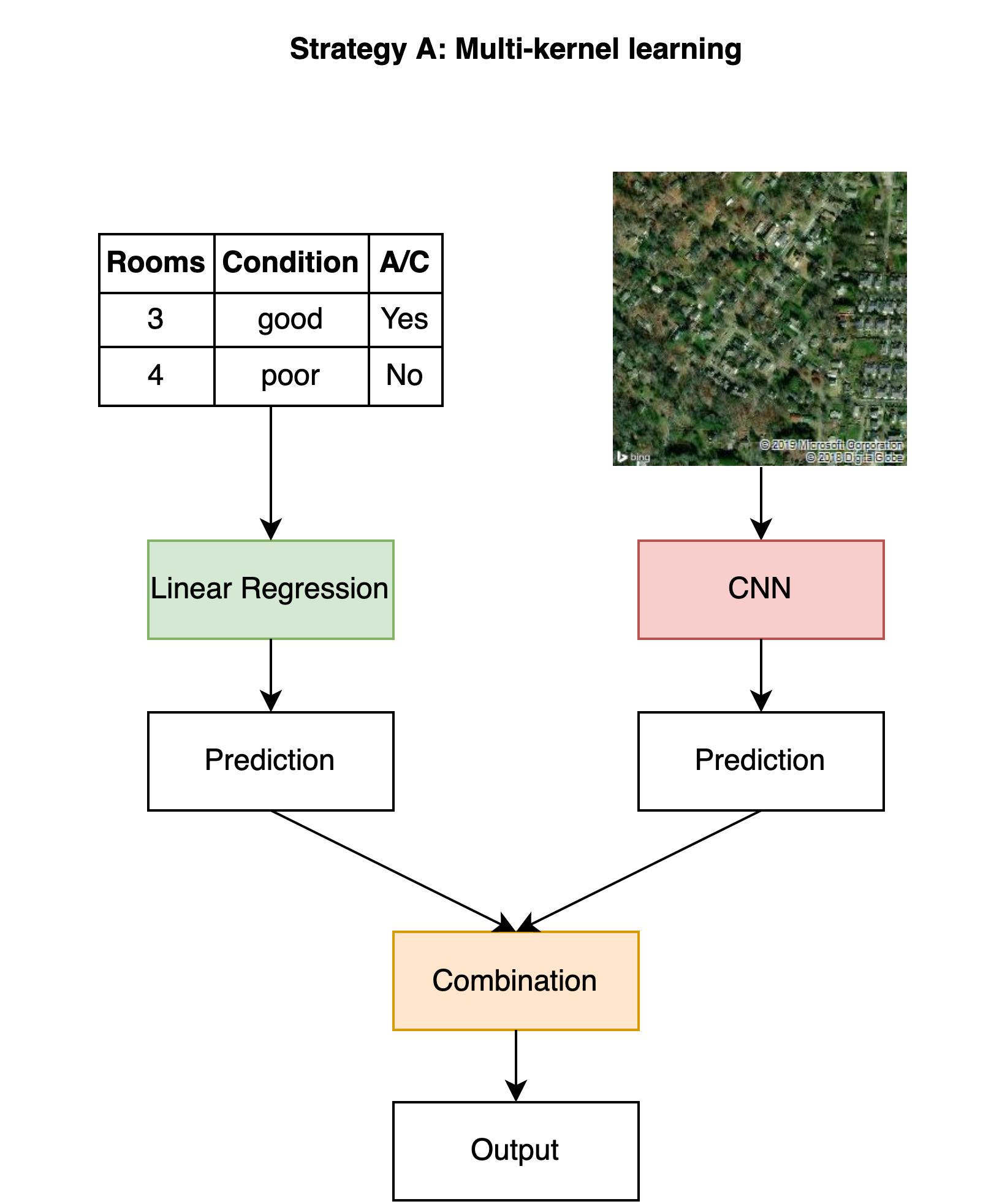} 
\caption{Strategy A of multi-kernel learning. Satellite image \textcopyright \space Microsoft (2019).}
\label{strategyA}
\end{figure}

Multi-kernel learning refers to the application of different learning algorithms to different views (see Figure \ref{strategyA}, Strategy A). Each kernel will focus on learning other aspects of the data and their results can then be combined linearly (with same or different weights), or non-linearly, or through kernel-boosting \cite{xu2013survey}.

\begin{figure}[t]
\centering
\includegraphics[width=0.9\columnwidth]{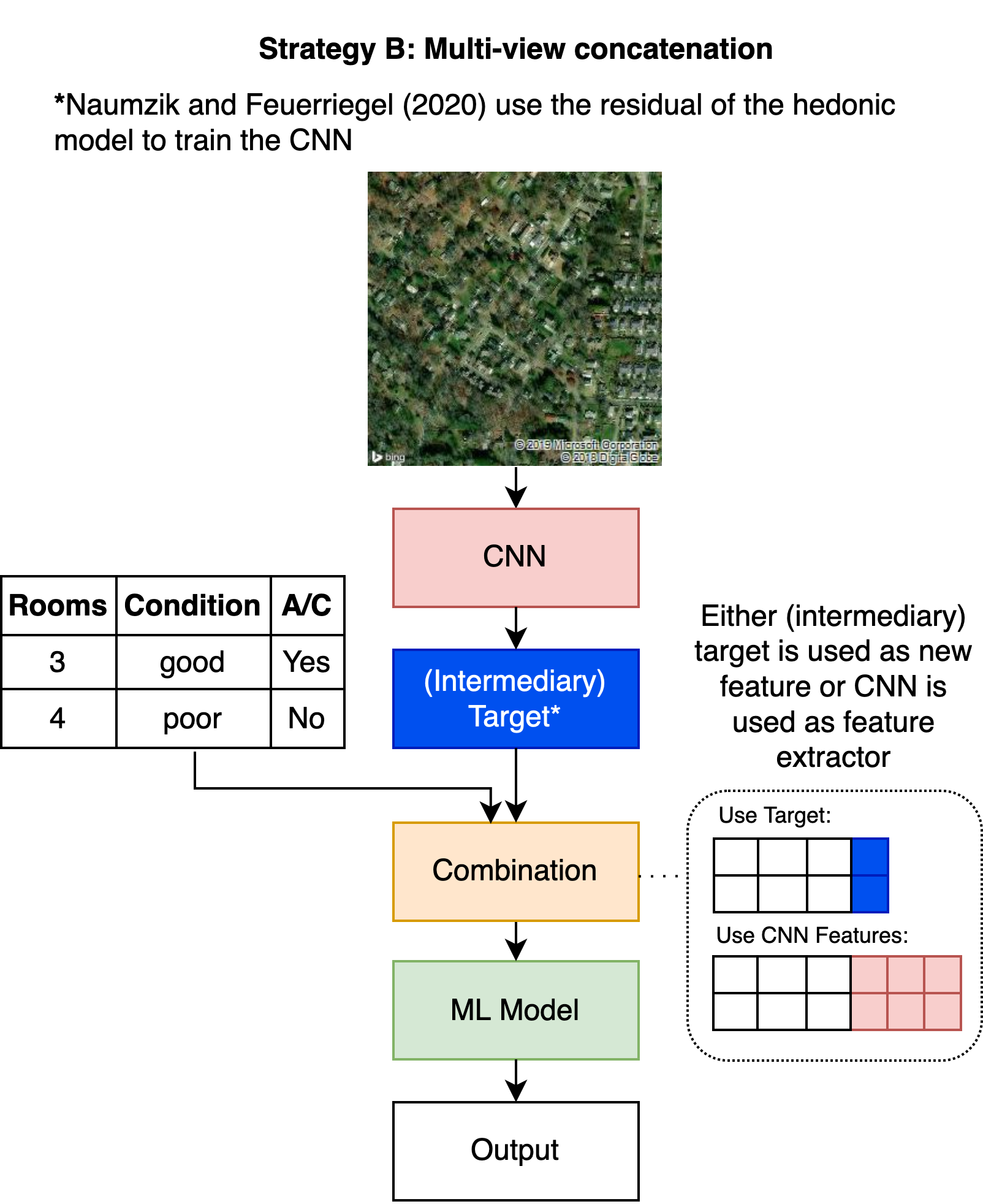} 
\caption{Strategy B of multi-view concatenation. Satellite image \textcopyright \space Microsoft (2019).}
\label{strategyB}
\end{figure}

An alternative idea is to simply concatenate features of the different views to create a single-view representation of the sources (see Figure \ref{strategyB}, Strategy B) \cite{zhao2017multi}. For a combination of the different data sources, the data structure needs to be harmonized before a matrix in form of the structured data (e.g., housing attributes) can be combined with a higher n-dimensional tensor of the images (e.g., satellite images). Therefore, the CNN either performs a classification of the image or a feature extraction by using a hidden layer of the neural network. For the former, the probabilities of each class can be represented in a tabular format and combined with the relational housing attributes. For the latter, the last fully connected layer before the (softmax) output in a neural network is often used. The gained features are stored in a matrix and joined with the relational housing data. However, especially when the combined view is high dimensional and contains only few observations, this strategy can easily lead to overfitting due to the curse of dimensionality \cite{xu2013survey}.

\begin{figure}[t]
\centering
\includegraphics[width=0.9\columnwidth]{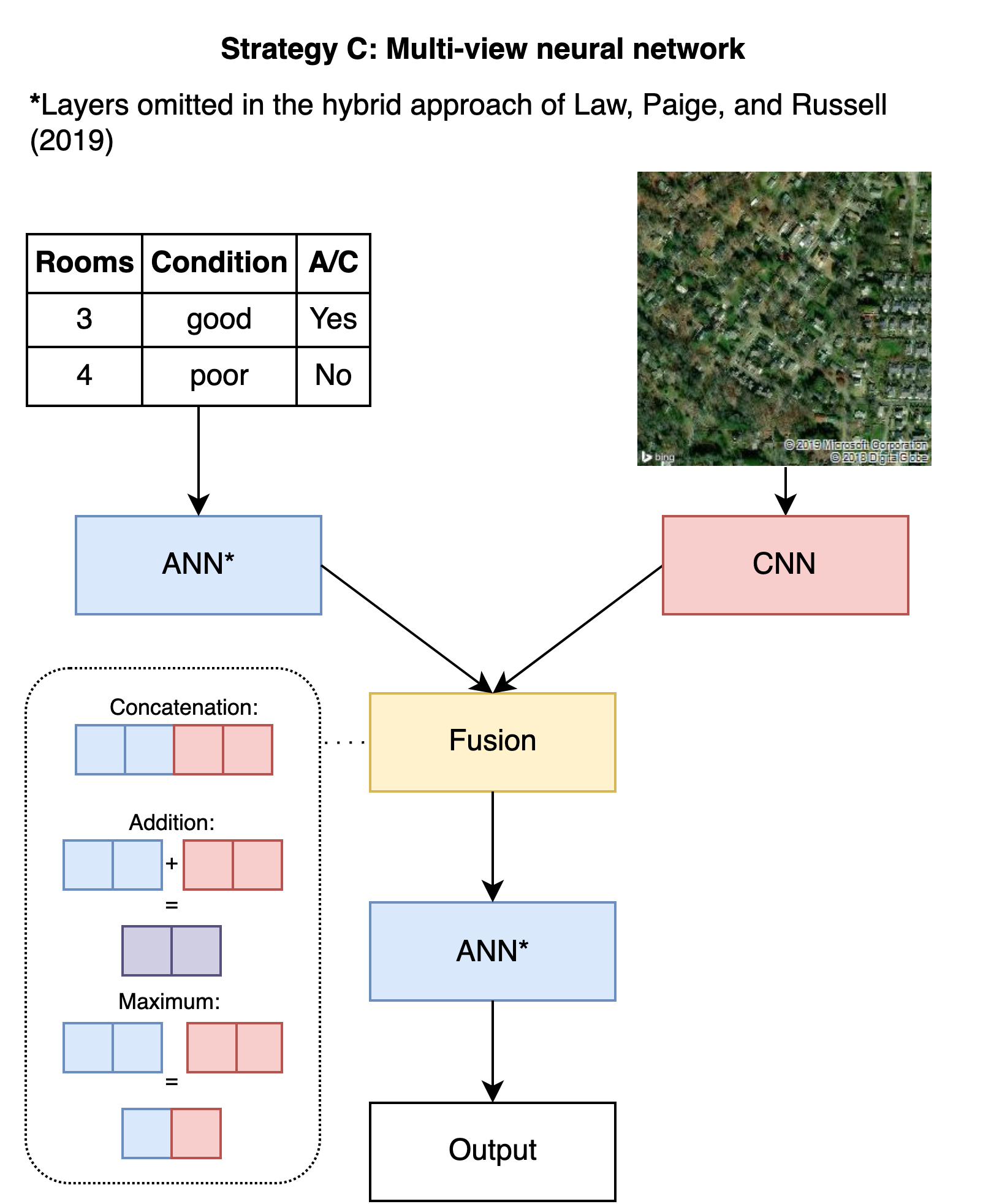} 
\caption{Strategy C of multi-view neural networks. Satellite image \textcopyright \space Microsoft (2019).}
\label{strategyC}
\end{figure}

The third class of strategies tries to learn a latent subspace of the views. Proposed methods are multi-view autoencoders and multi-view neural networks. Multi-view autoencoders, a self-supervised learning strategy, encode the information of the different views in a lower-dimensional subspace and try to rebuild it as accurately as possible by applying a decoder on the subspace. After training, the model's encoder part can extract the views' joint representation \cite{li2018survey}. Multi-view neural networks are similar to the autoencoder strategy, however, as they belong to supervised learning, these methods directly map the learned representation to a classification or regression output instead of reconstructing it \cite{xu2013survey} (see Figure \ref{strategyC}, Strategy C). For fusing the different views of the multi-view neural network, either concatenation, addition or maximum fusion can be used \cite{li2018survey}. 

Summarizing the differences between strategies A, B, and C, multi-view kernel learning focuses on an intelligent combination of algorithms and their predictions. Strategy B, the concatenation of views, is more data-focused. The overall goal is to concatenate features from different views before learning a predictive model. If the views have different representations, some of them have to be pre-processed (e.g., by using a CNN to extract features from images) before they can be joined. In strategy C, a multi-view neural network tries to learn a latent joint representation from multiple views at once.

\subsection{Multi-View Learning in Real Estate Appraisal}
In previous research on real estate appraisal, multi-view concatenation and multi-view neural networks were used \cite{liu2018learning, law2019take, bency2017beyond}.

In most cases where the authors have used multi-view concatenation, a non-linear machine learning algorithm (e.g., Random Forest, Support Vector Machine) was used to combine structured features and image features \cite{bency2017beyond, poursaeed2018vision, bessinger2016quantifying}. However, to extract the image features, authors have followed different approaches. For example, \citeauthor{bessinger2016quantifying} \citeyearpar{bessinger2016quantifying} used a CNN pre-trained on the places-365 data set predicting class probabilities representing the information extracted from the image. These additional features are used in combination to the classical housing attributes. Similarly, \citeauthor{poursaeed2018vision} \citeyearpar{poursaeed2018vision} trained a CNN to classify interior images into different luxury levels, which are used as additional features for the relational data. \citeauthor{bency2017beyond} \citeyearpar{bency2017beyond} followed a different strategy and did not use a model trained on a different data set for feature extraction, but trained their satellite-image based CNN on a binary intermediary target from the same data set, distinguishing the top and bottom 10\% of the prices. Instead of using the two classes as additional variables, they used the CNN's second last layer as a feature extractor for their downstream model.

The approach of \citeauthor{naumzik2020one} \citeyearpar{naumzik2020one} differs substantially from the other approaches of multi-view concatenation. To extract features from the images, the authors used the error term of the linear regression on the structured house features as an intermediary target for training the CNN. The final house price was then estimated based on another linear regression using the structured features and prediction from the CNN as inputs. This approach has some similarities to boosting, where one algorithm is trained on the other's error term to improve the overall performance. But instead of using just a single data source (as in boosting), each learner is trained on a different view. The strength of this model is its interpretability: The authors conclude that each standard deviation change in the interior image results in a price increase of 13.28\% \cite{naumzik2020one}.

Contrary to the multi-step strategies of multi-view kernel learning and multi-view concatenation, multi-view neural networks are a custom type of neural network that can handle multiple data input streams within one model (see Figure \ref{strategyC}). For each input stream, a particular branch adapted to the data type exists, e.g., a CNN for image data or an LSTM for textual data \cite{li2018survey}. Without intermediary steps like other target variables or the weighting of predictions, the model directly fits the multiple inputs to the target variable. Different architectures can be used, ranging from fully non-linear structures to semi-transparent designs \cite{law2019take}. The former can include multiple non-linear transformations in each of the neural network branches and additional non-linear layers after their fusion. Often, these types of architectures are used to maximize the predictive performance \cite{law2019take}. Nevertheless, these algorithms are not inherently interpretable, meaning that the effect of one input variable cannot be measured as for instance in a linear regression model. Driven by the design that predictions are combinations of multiple combinations of the input features, neural networks are typically categorized as black-box models, while semi-transparent architectures combine performance and interpretability. Hence, \citeauthor{law2019take} suggested a design without fully connected layers in the structured branch and after the concatenation. Additionally, the information of the image branch is compressed to a scalar. The resulting model could be described as a linear regression of the relational attributes and a new artificially created variable capturing the information of the image branch. Another option to increase transparency is the use of post-hoc interpretability methods, which can for example highlight important regions in the image \cite{kucklick2020location}.

After the technical description of the different strategies, we compare their benefits and disadvantages in detail. Combining views (see Figure \ref{strategyB}) is more favored in the related literature than the weighting of the different kernels. Strengths in the multi-kernel and multi-view concatenation approach lie in the low complexity of the training process. As the different data types are split over multiple models, which can be trained sequentially, only one algorithm at a time needs to be optimized. Besides lower resource requirements, it is reported that convergence is easier to reach compared to multi-view neural networks \cite{bency2017beyond}. This should result in a more stable learning process and consequently in a higher accuracy. Nevertheless, challenges for multi-view concatenation arise in choosing the intermediate target variable \cite{law2019take}. A suitable target needs to be selected so that the model can learn additional, complementary insights, for example, neighborhood types or luxury levels. In addition, labeled data is required but not always accessible. Thus, data preparation costs might be high when a supervised multi-view concatenation strategy is applied. For example, \citeauthor{poursaeed2018vision} \citeyearpar{poursaeed2018vision} labeled their own data set for measuring the luxury level. Another strategy might be the usage of a suitable pre-trained model, using extracted high-level image features, or training on residuals. The former strategy is promising when the image type is similar to the pre-trained model, e.g., exterior images and a place classifier. An alternative for using a pre-trained model or feature extracting, is training on residuals as it forces the image model to minimize the existing residuals of the housing attributes. In comparison, multi-view neural networks have the advantage that no intermediary target variable is required. Moreover, non-linear relationships of features and interactions between them can be modeled automatically \cite{law2019take}. However, downsides of multi-input neural networks are that their convergence is hard to achieve, and their black-box nature. To increase transparency and interpretability, \citeauthor{law2019take} suggested a hybrid architecture in which the authors leave out some non-linear layers. This approach is very similar to the suggested strategy of \citeauthor{naumzik2020one} \citeyearpar{naumzik2020one}, because both procedures focus on the interpretability. For both models, coefficients or even standard deviations, and significance levels can be extracted for a statistical interpretation of the results. Differences are that the hybrid model is a one-stage approach and thus does not require an intermediary target, while the boosting strategy of \citeauthor{naumzik2020one} is explicitly trained on the error term. In the next chapter, we compare the introduced strategies of multi-kernel learning, multi-view concatenation and multi-input neural networks on a shared dataset.

\section{Dataset and Modelling}

\begin{figure}[b]
    \centering
    \includegraphics[width=0.9\columnwidth]{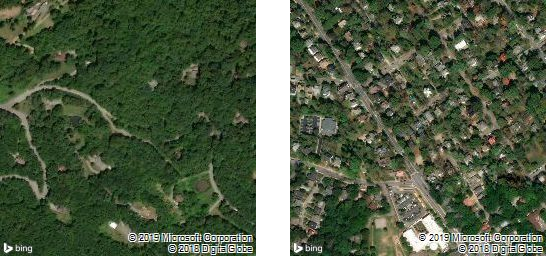}
    \caption{Examples of satellite images on zoom level 16. 
    Images \textcopyright \space Microsoft (2019). }
    \label{fig:example_img}
\end{figure}

\begin{table*}[t]
    \centering
    \begin{tabular}{|l|l|l|l|l|}
    \hline
        \textbf{Variable} & \textbf{Mean} & \textbf{Standard Deviation} & \textbf{Minimum} & \textbf{Maximum}  \\ \hline
        Square Feet & 3,061.81 & 1,303.81 & 1,022.00 & 7,352.00 \\
        Year Built & 1976 & 27.41 & 1790 & 2015 \\
        Full Bathrooms & 1.96 & 0.76 & 1.00 & 7.00 \\ 
        Half Bathrooms & 0.35 & 0.51 & 0.00 & 3.00 \\
        Bedrooms & 3.02 & 0.69 & 1.00 & 7.00 \\
        Acres & 0.76 & 0.76 & 0.06 & 4.88 \\
        Totalmarketvalue & 275,049.91 & 149,586.08 & 18,300.00 & 2,304,500.00 \\ \hline
    \end{tabular}
    \caption{Descriptive statistic of the numerical variables. The target variable is totalmarketvalue}
    \label{tab:summary_statistics}
\end{table*}

We used open data from 32,700 real estates in Asheville, NC for our experiments \cite{Ashville_1}. We combined different tables from the computer-assisted mass appraisal system, including details of the house and the lot. We focused on predicting prices for single-family homes, excluding condos and apartments, as they are difficult to separate from each other on satellite images. The dataset includes information about the size (number of rooms, bathrooms, lot size), year built, condition, quality, style (house type, roof style), and amenities (fireplace, indoor pool, garage spaces, air condition \& heating details) as well as the location. We include location fixed effects based on the city dummy variables in all models. The descriptive statistics are summarized in Table \ref{tab:summary_statistics}. The target variable is the totalmarketvalue and houses in the dataset have a mean value of approximately 275,000 USD. The property's address was used for obtaining the geographical coordinates via Bing Maps \cite{Bing}. In a second step, for each real estate we downloaded satellite images based on the property's geographical location. We decided to select a zoom level that allows to capture spatial features like parks and shops within walking distance (400-800m) to the real estate, as these features are known to influence prices \cite{noor2015sustainable, law2019take}. Hence, we chose zoom level 16 which depicts approximately 600m of surroundings in a 256 by 256 pixels image (Figure \ref{fig:example_img}). We evaluated our predictions using root-mean-squared error (RMSE) and mean absolute error (MAE). For a holistic view of the performance, we selected both measurements, as the MAE is robust to outliers, and the RMSE penalizes extreme values heavier. As a resampling strategy, we used a geographical out-of-sample dataset containing the two cities of Candler and Woodfin (n=1916) in the Asheville area, not represented in the training set. For training, we randomly split the remaining dataset into 80\% training set and 20\% validation set. Following the approach of \citeauthor{law2019take} \citeyearpar{law2019take}, we trained all models using the Adam optimizer for a maximum of 80 epochs. For the CNNs, we used the additive combination of RMSE and MAE as the loss function. We performed early stopping using the validation dataset to store the model with the best performance on the validation data. For better convergence of the neural network, we log-transformed the real-estate price and min-max normalized the relational and image data.

\begin{table*}[h]
    \centering
    \begin{tabular}{|l|l|l|l|}
    \hline
    \textbf{Model} & \textbf{Strategy} & \textbf{Target variable}& \textbf{Combination method} \\
    \hline
      Model 1  & A: Multi-kernel learning & Price & Linear weighting \\
      Model 2 & B: Multi-view concatenation by feature extraction & Price & Concatenation \\
      Model 3 & B*: Multi-view concatenation by boosting & Error term & Boosting \\
      Model 4 & C: Hybrid multi-view neural network & Price & Fusion by concatenation \\
      Model 5 & C*: Multi-view neural network & Price & Fusion by concatenation \\
    \hline
    \end{tabular}
    \caption{Summary of the different models compared.}
    \label{tab:trained_models}
\end{table*}

The tested models are summarized in Table \ref{tab:trained_models}. We built these different models based on the strategies depicted in Figures \ref{strategyA}, \ref{strategyB}, and \ref{strategyC}. Model 1 is a multi-kernel model trained with the price as target variable. One kernel uses hard housing attributes while the other kernel focuses on learning latent signals from the satellite images. Both models are weighted equally and linearly \cite{xu2013survey}. The architecture of model 2 is based on the concatenation of different views \cite{poursaeed2018vision, bency2017beyond, bessinger2016quantifying}. It uses features extracted from a CNN and fuses them with the house's relational data in a Random Forest to compute a final price estimate. Model 3 follows the approach of \citeauthor{naumzik2020one} \citeyearpar{naumzik2020one}, where a hedonic pricing model based on a linear regression is boosted with the error term prediction of a CNN. For the models 1 to 3, we used ResNet50 \cite{he2015delving} for the CNN as the basic architecture, followed by an additional fully connected layer with 128 neurons before the output. One problem that can occur when using deep neural networks are vanishing gradients. The ResNet architecture based on skip-connections is less prone to this phenomenon than other architectures. Finally, models 4 and 5 are multi-view neural networks \cite{li2018survey}. Both are based on the approach of \citeauthor{law2019take} \citeyearpar{law2019take}. Instead of following the network architecture of \citeauthor{law2019take}  \citeyearpar{law2019take} using VGG-16 for the image branch \cite{simonyan2014very},  we again used the ResNet50 architecture including the additional dense layer. For model 5, we used one fully connected layer with 64 neurons and a relu activation in the housing data branch, and another fully connected layer with 64 neurons and a relu activation after the concatenation of the branches. While model 4 is semi-transparent, model 5 is a black-box model. In line with previous real estate appraisal research, we used a standard linear regression model as a baseline without modeling interaction terms  \cite{limsombunchai2004house}. The results of our experiments are reported in Table \ref{tab:results_performance}.

\section{Results and Discussion}
We separate the following section into two. First, we evaluate the different strategies A to C from a metrical perspective on the MAE and RMSE. Secondly, we show and compare model 3 and 4 concerning the interpretability of coefficients. 

\begin{table}[b]
    \centering
    \begin{tabular}{|l|l|l|}
    \hline
        \textbf{Model} & \textbf{MAE} & \textbf{RMSE}  \\ \hline
        Baseline & 40,303  & 71,518 \\
       Model 1 & 49,019  &  78,983  \\
        Model 2 & 38,395 & 61,663  \\ 
        Model 3 &  43,173  & 68,362 \\
        Model 4 & 37,225 & 61,429 \\ 
        Model 5 & 34,890 & 56,099 \\ 
        \hline
   \end{tabular}
   \caption{Performance of the different models. Before evaluation, the log-transformation is inversed to gain better interpretability.}
    \label{tab:results_performance}
\end{table}

\subsection{Predictive Accuracy}
The results summarized in Table \ref{tab:results_performance} suggest that adding satellite images to standard structured features improves the predictive accuracy of real estate appraisal models. The different strategies lead to a reduction of up to 5,413 USD in MAE, which equals a performance improvement of 13.4\%. As these improvements are similar in magnitude to those in related research, the results provide further empirical evidence that satellite images are of importance for real estate appraisal \cite{law2019take, bency2017beyond}. However, the performance gains vary substantially between the different multi-view learning strategies. Model 1, applying a multi-kernel learning strategy, performs significantly worse than the baseline. This may be explained by the different predictive powers of the two kernels. While the linear regression had a MAE of 40,303 USD, the CNN had a MAE of 73,023 USD. This suggests that satellite images alone are unable to precisely predict real estate value and that weighting the prediction does not improve the estimation. In contrast, a multi-view concatenation strategy (models 2 and 3) leads to performance gains of up to 4.7\% in MAE compared to the baseline. It seems that the modeled non-linear combination of structured house attributes and extracted image features in model 2 is beneficial for estimating real estate prices. Model 3, applying the boosting strategy of \citeauthor{naumzik2020one} \citeyearpar{naumzik2020one}, had a lower RMSE but higher MAE compared to the baseline. We can think of two explanations for this phenomenon. First, the satellite image view used was not entirely complementary (in other words: overlapping) to the housing attributes, which also contain location information. In the work of \citeauthor{naumzik2020one} \citeyearpar{naumzik2020one}, visual aesthetics extracted from the interior images might have improved predictive performance, because these features were not captured by the relational attributes. Second, using not only an independently and identically distributed hold-out set, but an out-of-sample test set further increases the required maturity level of the algorithm. On the in-sample validation dataset, the performance improved on RMSE and MAE. Nevertheless, on the out-of-sample predictions, only a lower RMSE could be reported (Table \ref{tab:results_performance}). As the RMSE penalizes larger errors stronger, it seems that extreme deviations are reduced. 
Though, the higher MAE could be caused by noise introduced by CNN's prediction of the residual. Finally, models 4 and 5, both based on the multi-view neural network strategy, performed best by far. Model 4, the interpretable hybrid model provided 7.6\% more accurate predictions in terms of MAE. Model 5, the more complex black-box model, outperformed the baseline by 13.4\% in MAE. In the next section we show the possibilities to interpret models 3 and 4.

\subsection{Interpretability of Multi-View Real Estate Appraisal Models}

\begin{table}[b]
    \centering
    \begin{tabular}{|l|l|l|}
    \hline
        \textbf{Variable} & \textbf{Model 3} & \textbf{Model 4} \\ \hline
         Constant & 11.07 & 0.27 \\
         Square Feet  & 0.98  & 1.00 \\
         Year Built  & 0.34 & 0.53 \\
         Half Bathrooms   & 0.12 & 0.12 \\
         Fireplace & 0.05 & 0.05 \\
         Style 1.5 Conventional & 0.05 & 0.04 \\
         Locality Asheville & 0.17 & 0.11 \\
         Condition Unsound & -0.23 & -0.33 \\
         \hline
         Satellite Image  & 1.05 &  0.41 \\
         \hline
    \end{tabular}
    \caption{Selective set of coefficients of model 3 and 4}
    \label{tab:coefficients}
\end{table}

\begin{table*}[t]
    \centering
    \begin{tabular}{|l|l|l|l|}
    \hline
        \textbf{Strategy} & \textbf{Benefit} & \textbf{Disadvantage} & \textbf{Suggested use} \\ \hline
  A & Low training complexity & Lower accuracy & Advanced baseline model \\ \hline
         B & Sequential training process & Difficult target variable selection & Alternative for strategy C \\ \hline
 B* & Statistical interpretation & Complementary views required & Research purposes \\ \hline
         C & Interpretability & High complexity of training & Multi-use \\  \hline
 C* & Performance & High complexity of training & Predictive model \\
         \hline
    \end{tabular}
    \caption{Summarization of benefit and disadvantage of the different strategies}
   \label{tab:strength}
\end{table*}

We report a selective set of coefficients of model 3 and model 4 in Table \ref{tab:coefficients}. While the constants differ between the models, the estimated effect of the size measured in square feet, the amenity fireplace or the location in Asheville have approximately the same size in both models. For example, the house price increases by approximately 5\% when the house has a fireplace, ceteris paribus. As the continuous variables are min-max normalized, inferential statements cannot be made. Nevertheless, the sign and size of the coefficients show the direction and strength of impact on the house price. For example, the condition unsound is associated with a negative price influence, while having additional half-bathrooms adds value to the real estate. From interpreting the coefficients magnitude, we can conclude that location aspects captured by the satellite image have a strong influence on the appraisal value, as the satellite image is among the top-3 largest coefficients for model 3 and top-15 for model 4. The difference between the magnitude of the image data coefficient of model 3 and 4 can be explained by the different scales of the variables. While model 3 includes the impact of the satellite image by the predicted residual and is thus measured in log-USD, model 4 learned the impact of the image implicitly. When the predicted residual changes by 1\%, the house price changes by 1.05\% (model 3). Furthermore, model 3 has the advantage that besides the coefficients, also standard errors, t-values, and significance levels can be extracted from the linear regression model. Concluding, the derived coefficients and statistical measures can enhance the model understanding and transparency.

In summary, we can conclude the following lessons learned from our experiments (Table \ref{tab:strength}): Models using multi-kernel learning (Strategy A) are easy to train, however their performance seems to depend strongly on the weighting of the kernels. We suggest using these models as an advanced baseline, in addition to classic hedonic pricing models. Furthermore, models based on multi-view concatenation (Strategy B) seem to reliably increase predictive performance. This strategy is easier to optimize compared to multi-view neural networks, however, selecting the right (intermediate) target seems to be essential for successful learning. Moreover, it seems that the boosting approach of \citeauthor{naumzik2020one} \citeyearpar{naumzik2020one} (Strategy B*) is more beneficial when the additional view is largely complementary to the other view. Nevertheless, the strength of this approach lays in interpretable coefficients and a statistical measurement of the effects. Consequently, this strategy seems very suitable for research contexts. From our experience, multi-view neural networks (Strategy C and C*) perform best. It seems that learning a latent subspace leads to an effective feature representation even if the multiple views overlap in some features. In our experiment, the location of the house is, in addition to the satellite image, also partly captured by city dummy variables. On the downside, according to \citeauthor{bency2017beyond} \citeyearpar{bency2017beyond}, it can become difficult to achieve convergence of the neural network model, which raises the complexity of training. In addition, model 5 (Strategy C*) comes as a black-box model. To mitigate this weakness, a semi-transparent multi-neural network as suggested by \citeauthor{law2019take} \citeyearpar{law2019take} can enhance interpretability. Nonetheless, it comes with an interpretability-accuracy trade-off, as model 4 has weaker predictive performance than model 5. We suggest using the multi-view neural networks for predictive models in applications where explainability is not a key objective.

\section{Conclusion}
Of course, our work is not without limitations. We have not yet investigated the interrelations between the structured housing attributes and the satellite images. Future research could, for example, analyze which information is overlapping between views and which information can only be captured by one or the other view. Another limitation is related to our data source. We use only a single dataset of Asheville, NC to assess the effect of the learning strategy on predictive performance. As related literature typically refers to neural network architecture search, future research should perform replication and ablation studies to examine, if results can be reproduced across datasets, image types and domains (beyond housing).

Despite these limitations, the following implications for research and practice can be derived from our findings. Satellite images can clearly improve the accuracy of computer-assisted mass appraisal. Our results indicate that the MAE can be reduced by up to 13\%, depending on the chosen multi-view learning strategy. Therefore, banks and lenders should consider using visual data to improve their real estate appraisal estimates. Moreover, different techniques match different purposes and user groups. Researchers interested in a statistical interpretation of the results might find the boosting strategy of model 3 more appealing. Practitioners mainly interested in predictive accuracy might prefer model 5, the multi-view neural network.

\bibliography{references}

\end{document}